\documentclass{article}
\usepackage{spconf,amsmath,graphicx}
\usepackage{subcaption}     
\usepackage{booktabs}
\usepackage{multirow}

\title{Audio-Visual Event Recognition through the lens of Adversary}
\name{Juncheng B Li, Kaixin Ma, Shuhui Qu, Po-Yao Huang, Florian Metze}
\address{junchenl, kaixinm, poyaoh, fmetze@cs.cmu.edu; shuhuiq@stanford.edu}
\begin{document}
\ninept
\maketitle
\begin{abstract}
As audio/visual classification models are widely deployed for sensitive tasks like content filtering at scale, it is critical to understand their robustness along with improving the accuracy. This work aims to study several key questions related to multimodal learning through the lens of adversarial noises: 1) The trade-off between early/middle/late fusion affecting its robustness and accuracy 2) How does different frequency/time domain features contribute to the robustness? 3) How does different neural modules contribute against the adversarial noise? In our experiment, we construct adversarial examples to attack state-of-the-art neural models trained on Google AudioSet.\cite{gemmeke2017audio}\footnote{Audioset is the largest available weakly-labeled audio dataset.} We compare how much attack potency in terms of adversarial perturbation of size $\epsilon$ using different $L_p$ norms we would need to ``deactivate" the victim model. Using adversarial noise to ablate multimodal models, we are able to provide insights into what is the best potential fusion strategy to balance the model parameters/accuracy and robustness trade-off, and distinguish the robust features versus the non-robust features that various neural networks model tend to learn.\footnote{Preprint Under review.}
\end{abstract}
%
%
\section{Introduction}
\label{sec:intro}

Increasingly, audio/visual event recognition (AER) is gaining importance as its application in content filtering and and multimedia surveillance is soaring.~\cite{huang2018multimodal} These days, uploading a clip of video or audio to social media platforms such as Facebook or Youtube would involve going through its internal algorithmic content filtering mechanism in case of potential policy-violating contents. This requires such systems to be both accurate and robust in order to sniff out(classify) malicious content, which could potentially reduce the need for human intervention. 

From a multimodal learning perspective, the AER task is a perfect angle to understand the interactions between the audio and video modalities without worrying about the dependencies or senses in spoken language, as there is no language model involved~\cite{wang2018polyphonic}. In this work, we base our experiments on the task of audio tagging, which aims to characterize the acoustic event of an audio stream by selecting a semantic label for it, and we perform analysis on the largest available dataset: Google AudioSet~\cite{gemmeke2017audio}.
The benchmark on the AudioSet first introduced by~\cite{vggish} has been constantly refreshing and growing from a mean average precision (mAP) 31\% to 43.1\% as reported by~\cite{kong2019panns,wang2018polyphonic,wang2019comparison,kong2018audio}.

Downstream applications of such AER models on home-security cameras detecting events such as glass-breaks, dog-barks or screaming have a strict requirement for both accuracy and robustness to achieve practicality.
Despite the sheer volume of existing works on improving the benchmark of the AER task~\cite{Mesaros2016_MDPI, kong2019panns} from the audio/visual research community, the reasoning for accuracy improvements and robustness of the models is still lacking. Most existing works on understanding audio classification either performed post-hoc analysis or concluded some intuitions from observations~\cite{li2017comparison, kong2018audio, alwassel2019self}. Motivated by this, this work aim to explain neural networks' behavior on audio/visual data from an \textit{adversarial perspective} to dissect the multi-modal learning process. 

We follow the framework: Re-implement/improve STOA models $\rightarrow$ Generating adversaries $\rightarrow$ Identify what breaks the model $\rightarrow$ Point out what makes the networks more robust $\rightarrow$ Understand how to distinguish robust features/non-robust features.\\

\section{Background and Related Works}

\textbf{Audio/Visual Event Classification: } 
Recently, several neural network models have been applied to the AER task and demonstrated their improvement in performance. \cite{xu2018large, wang2019comparison, Yu2018MultilevelAM, Kong2019WeaklyLA} showcased the effectiveness of using the Convolutional Recurrent Network (CRNN) architecture which has become the main-stream architecture for audio classification. \cite{Ford2019} demonstrated that pure convolutional neural network (ResNet) without RNNs could also work well on audio. Self-attentioned Network~\cite{Vaswani:2017:AYN:3295222.3295349} has recently become very popular and has also been applied to this task.~\cite{9053609} Other work such as~\cite{wang2019makes} has incorporated visual modalities to boost the performance.
In this paper, we provide a comprehensive comparison between our model with CRNN and ResNet on the AER task from the adversarial perspective. 

 \noindent \textbf{Adversarial Examples: } Fully explaining behaviors of the non-convex non-linear neural networks is notoriously difficult if not a mission impossible. We do not believe conventional wisdom could provide a ``Panacea" at the moment. Inspired by~\cite{ilyas2019adversarial} which demonstrated that adversarial examples are actually features in the vision domain, we envision that computing adversarial examples might help us identify what features different neural networks are learning from the audio-visual data, and whether they learn robust features. Adversarial examples were widely recognized to be a security concern for machine learning, and they are recently demonstrated to be equally effective in the audio domain~\cite{li2019adversarial}, in the meantime, there were attempts to study them in audio tasks~\cite{subramanian2019,du2020sirenattack}. We are not only focusing on the security aspect of adversarial examples in this work, we also believe by going in the opposite direction of gradient descent (adversarial attack) we could potentially gain more insights.

The main contributions of our paper can be summarized as:
\begin{enumerate}
    \item We introduce a convolutional self-attention network for audio/event event classification task, using visual features together with audio features, we can achieves 44.1 mAP on the Google Audioset, which further improves the SOTA benchmark on this task.
    \item Leveraging adversarial examples, we showcase what are the robust features in frequency and time domain;
    \item We provide a detailed analysis on the proposed multi-modal architecture through adversarial examples, and address the trade-off between early, middle and late fusion on adversarial robustness;
\end{enumerate}

\section{Convolutional Self-attention Network (CSN)}


\begin{figure}
    \centering
    \includegraphics[scale = 0.3]{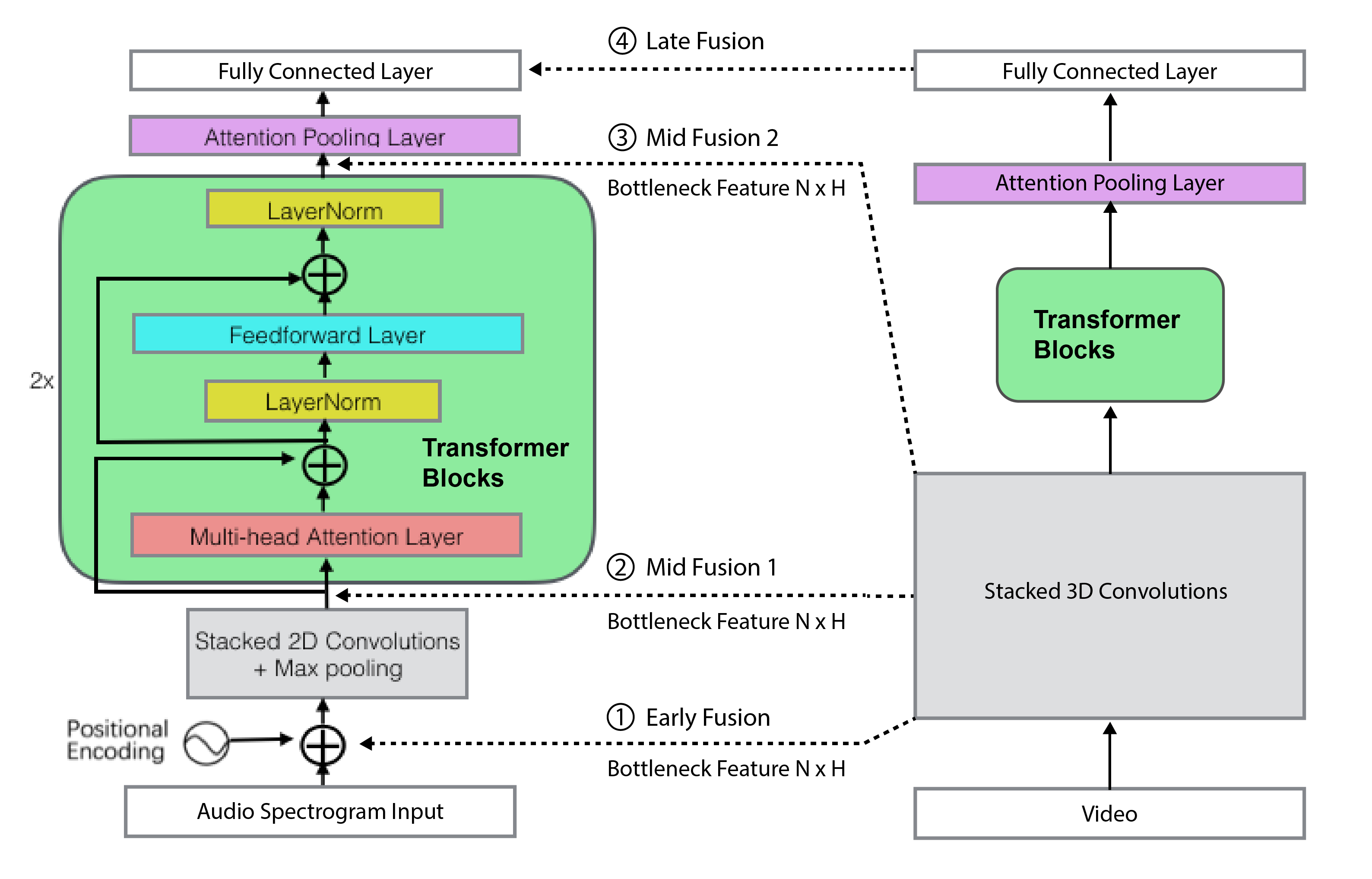}
    \caption{The overall architecture of Convolutional Self-attention Network and the different multimodal fusion strategies studied.}
    \label{fig:CSN}
\end{figure}

\subsection{Audio Encoding Network}

We first describe the audio encoding part of CSN. Following TALNet architecture proposed by \cite{wang2018polyphonic}, we employ stacked convolutions and pooling layers to first extract high-dimensional features from raw frames. In particular, the audio encoder consists of 10 convolution layers of 3x3 and 5 pooling layers are insert after every 2 convolution layers, which reduce the frame rate from 40 Hz to 10 Hz. The outputs of the convolution encoder are fed into 2 transformers blocks to further model the global interaction among frames. Each transformer block consists of 1 layer of multi-head scaled dot-product attention and 2 feed-forward layers which are connected by residual connections. Finally, we use attention pooling \cite{wang2018polyphonic} to make final predictions. The audio encoder of CSN is depicted in the left branch of Fig \ref{fig:CSN}. 


\subsection{Video Encoding Network (3D-CNN)}
\label{sec:ven}
For encoding and representing videos, our model employs the R(2+1)D block similar to~\cite{tran2018closer} which decomposes the 3D (spatial-temporal) CNN into a spatial 2D convolution followed by a temporal 1D convolution. 
Specifically, let $h, w, k$ denotes the height and width of $k$ frames in a sliding window. For a video clip with $M$ frames, there are $N = \lceil{M / k}\rceil$ windows in it, and the size of each window is denoted as $[k,w,h]$. 
For each $k$ frame window, our 3D CNN employs a sequence of $[1,w,h]$ 2D convolutional kernel followed by a $[k,1,1]$ temporal convolutional kernel
to encode the video clip into a size $H$ vector.
For the windows in the video, we utilize a 1D temporal CNN and a linear layer to extend its temporal receptive field and map the visual context for multi-modal fusion with the audio model.
Essentially, the 3D CNN encodes the video into $\mathbf{v} \in \mathcal{R}^{H \times N}$, where $H$ is the embedding size for multimodal fusion. 
For late-fusion, we use additional 2 layers of transformers and an attention pooling layer for prediction. The visual modules are shown on the right side of Fig~\ref{fig:CSN}.

\subsection{Multimodal Fusion}
In order to understand the trade-offs between early, middle and late fusion, we performed 4 different types of fusions in our study. As is shown in right side of Fig.~\ref{fig:CSN}, we analyze the effects of fusing video bottle-neck features with the audio pipeline at different stages. 
Specifically, we investigated late fusion, two levels of middle fusion, and early fusion.
For late fusion (maximum model parameters), we aggregate the output predictions from the full audio and visual models. For early fusion (minimum model parameters), we concatenate the visual bottleneck feature with the spectrogram. In middle fusion, we concatenate the visual feature before and after the transformer blocks with the audio bottleneck features.
 
As is pointed by~\cite{wang2019makes}, in training multi-modal models, the different modalities tend to overfit and generalize at different rates. Thus, we normalize and balance the gradients of the two modalities before we transform and concatenate the latent dimensions in the fusion process.

\section{Adversarial Perturbations}
\subsection{Projected Gradient Descent}

Originally, single-modal AER models can be formulated as a minimization of $\mathbf{E}_{x,y\sim D}[L(f(x),y)]$ where $L$ is the loss function, $f$ is the classifier mapping from input $x$ to label $y$, and $D$ is the data distribution. We evaluate the quality of our classifier based on the loss, and a smaller loss usually indicates a better classifier.
In this work, since we are using adversarial noises to deactivate the models, we form $\underset{\delta}{\max}[\mathbf{E}_{x,y\sim D}[L(f(x'),y)]]$, where $x' = x + \delta$ is our perturbed audio input. It is common to define a perturbation set $C(x)$ that constrains $x'$, such that the maximization problem becomes $\underset{x' \in C(x)}{\max}[\mathbf{E}_{x,y\sim D}[L(f(x'),y)]]$. The set $C(x)$ is usually defined as a ball of small radius of the perturbation size $\epsilon$ (of either $\ell_\infty, \ell_2 \,$ or $\, \ell_1$) around $x$. 

To solve such a constrained optimization problem, one of the most common methods utilized to circumvent the non-exact-solution issue is the Projected Gradient Descent (PGD) method:
\begin{equation}
    \delta = \mathcal{P}_{\epsilon} \left(\delta - \alpha \frac{\nabla_{\delta}L(f( x +\delta),y)}{\Vert \nabla_{\delta}L(f( x +\delta),y)\Vert_p}\right)
\end{equation}
where $\mathcal{P}_{\epsilon}$ is the projection onto the $\ell_p$ ball of radius $\epsilon$, and $\alpha$ is the gradient step size.
To practically implement gradient descent methods, we use very small step size and iterate it by $\epsilon / \alpha$ steps. 

In this work, we are particularly interested in measuring the attack potency through the benchmarks including $\epsilon$ in the $\ell_p$ norm ball, and step size $\alpha$. Since our main focus is to maximally exploit the model's universal weakness, we only discuss untargeted attacks that drift the model's prediction away from the true label $y$. We do not discuss targeted attacks where we have to minimize the loss towards $y_{target}$ in the mean time, since doing this will inevitably further constrain our optimization and make our attack less potent. Note that our $\delta$ is a universal perturbation trained on the entire training set, and it is tested against the entire evaluation set to measure its attack success rate.

\subsection{Multi-Modal Adversarial Perturbations}
In a multi-modal setting, the loss formulation is $L_{multi} = L(f (x_{m_1} \oplus x_{m_2} \oplus \cdots \oplus x_{m_k}), y)$, where $x_{m_k}$ denotes inputs from multiple modalities. In this work, we are dealing with 2 modalities: $x_{audio}$ and $x_{video}$ We study the adversarial perturbation computed against the audio input $\delta_{audio}$. Our optimization goal is: $\underset{\delta_{audio}}{\max}[\mathbf{E}_{x,y\sim D}[L(f((x_{audio} + \delta_{audio}) \oplus x_{video}),y)]]$ , and thus, our PGD step is: 
\begin{equation}
\begin{aligned}
    & \delta_{audio} = \\
    & \mathcal{P}_{\epsilon} \left(\delta_{audio} - \alpha \frac{\nabla_{\delta_{audio}}L(f(( x_{audio} + \delta_{audio}) \oplus x_{video}),y)}{\Vert \nabla_{\delta_{audio}}L(f((x_{audio} + \delta_{audio}) \oplus x_{video}),y)\Vert_p}\right)
    \end{aligned}
\end{equation}
Since we focus more on the robustness of the audio modality in this work, we will study the effect of $\delta_{video}$ in a future work, and we expect the same formulation applies to the visual modality. 

\section{Experiments}
\subsection{Dataset}
We downloaded the entire audio and visual track of Google Audio Set~\cite{gemmeke2017audio}\footnote{Unfortunately, some links are broken, we were able to download 98\% of the 2 million clips of Audioset} which contains 2~million 10-second YouTube video clips, summing up to 5,800 hours. AudioSet is annotated with 527 types of sound events occurring in the clips, but they are weakly labeled in the sense that no information about the time span (onset and offset) of the events is annotated in each 10-second excerpt. We train and test the models according to the train and test split described in~\cite{vggish}.
The input for the audio branch are matrices of Mel filter bank features, which have 400 frames(time domain) and 64 Mel-filter (frequency domain) bins. For the visual branch, we employ a 3D CNN backbone to encode the spatial-temporal context in videos as is mentioned in Section~\ref{sec:ven}.

\subsection{Training Details}
\textbf{3D-CNN}: Our 3D CNN video encoding backbone is initialized with the network pre-trained on IG65M~\cite{GhadiyaramTM19}. For training on the videos in Google Audio Set, we freeze the clip-level 3D-CNN backbone and fine-tune the video-level 1D CNN and the transformer layer.

\noindent \textbf{CSN}: We used batch size of 300 and train the model for 25K steps. The model is optimized with Adam with learning rate of 4e-4. We used dropout rate of 0.75 for transformer layers and we found this critical to achieve the best performance. Also we did not use positional encoding as we found that model would yield better performance without it. 


\noindent \textbf{CRNN and ResNet}: We reimplemented CRNN according to TALNet~\cite{wang2018polyphonic}, and ResNet following the same setup as~\cite{Ford2019} as baselines for comparisons in Section~\ref{sec:archi}.


\section{Results and Discussion}

\subsection{Adversarial Examples against Fusion in different stages}
As we can see from Table~\ref{tab:fusion}, late fusion prevails in terms of performance with and without the presence of the adversarial noise computed in the same way (full-frequency and time domain, the same size as input). However, late fusion is the most expensive model as it requires 2 full-sized branches of audio and visual models with a huge amount of  parameters. Evidently, we can observe the trend that the deeper level the fusion is, the more robust the model would be against the adversary. 


\begin{table}[h]
\centering
\begin{tabular}{@{}ll | rrr@{}}
\toprule
Models  & Attack   &{\bf mAP} & {\bf AUC} & {\bf d-prime}  \\ \midrule
Early Fusion  & No  & 0.41 & 0.920 & 2.360     \\
Early Fusion  & Yes & 0.07  & 0.811  & 1.172     \\
Mid Fusion 1  & No  & 0.41 & 0.931 & 2.460    \\
Mid Fusion 1  & Yes & 0.19    & 0.832   & 1.268\\
Mid Fusion 2  & No  & 0.44 & 0.970 & 2.660    \\
Mid Fusion 2  & Yes & 0.21    & 0.865   & 1.378  \\
Late Fusion   & No  & \bf{0.44} & 0.969 & 2.631  \\
Late Fusion   & Yes & \bf{0.27}    & 0.910   & 1.871 \\\bottomrule
\end{tabular}
        \caption{Performance of our best performing CSN models with different multimodal fusion strategies shown in Fig~\ref{fig:CSN}, and their performance against the same strength of adversarial perturbation ($\epsilon = 0.3$, $\ell_{2}$ norm). Here, mAP is the mean average precision, AUC is the area under the false positive rate and true positive rate (recall) which reflects the influence of the true negatives. The d-prime can be calculated from AUC~\cite{gemmeke2017audio}.}
        \label{tab:fusion}
\vspace{-0.5cm}
\end{table}

\subsection{Constrained Temporal/Frequency Adversarial Examples}

SpecAugment~\cite{Park2019SpecAugmentAS} introduced a simple but effective technique for data augmentation in speech recognition task, where they mask out certain frequency band or time steps in the original spectrogram input. Inspired by this, we modify the spectrograms to construct adversarial examples. In particular, we distinguish frequency bands that are robust features versus those that are non-robust. Our results also partially explains why \cite{Park2019SpecAugmentAS}'s approach achieved significant performance improvement, that is by masking out non-robust frequency bands, the model is forced to learn more robust features that are going to generalize better in test setting.  

\begin{figure}[t!h!]
\begin{subfigure}[h]{0.243\columnwidth}
\raisebox{-\height}{\includegraphics[width=\columnwidth]{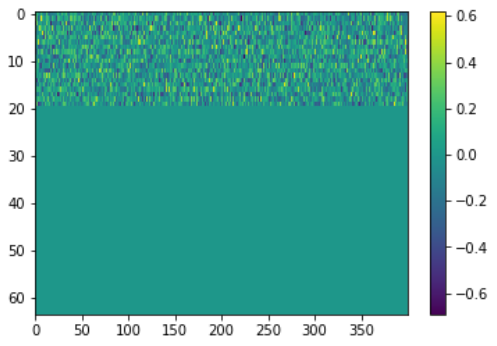}}
\centering
\caption{Low Freq Noise(0-20)}
\end{subfigure}
\hfill
\begin{subfigure}[h]{0.24\columnwidth}
\raisebox{-\height}{\includegraphics[width=\columnwidth]{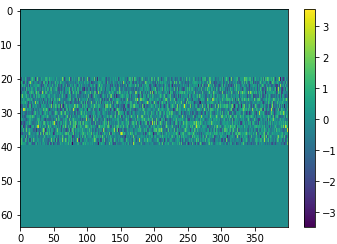}}
\centering
\caption{Higher Freq Noise (20-40)}
\end{subfigure}
\begin{subfigure}[h]{0.24\columnwidth}
\raisebox{-\height}{\includegraphics[width=\columnwidth]{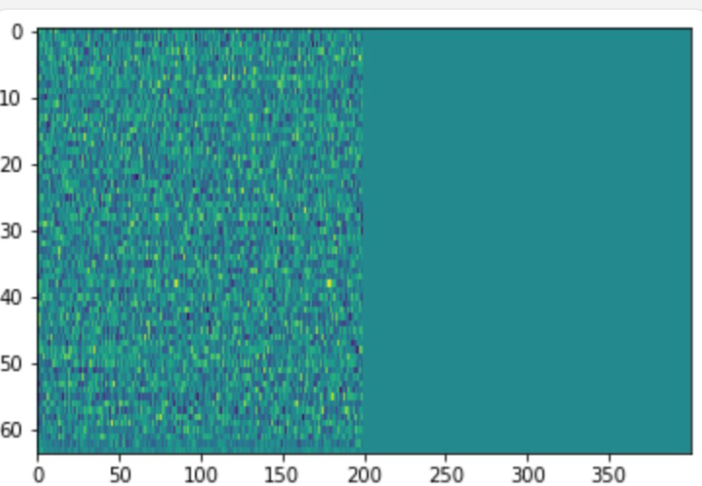}}
\centering
\caption{Temporal Noise (0-200dim)}
\end{subfigure}
\hfill
\begin{subfigure}[h]{0.24\columnwidth}
\raisebox{-\height}{\includegraphics[width=\columnwidth]{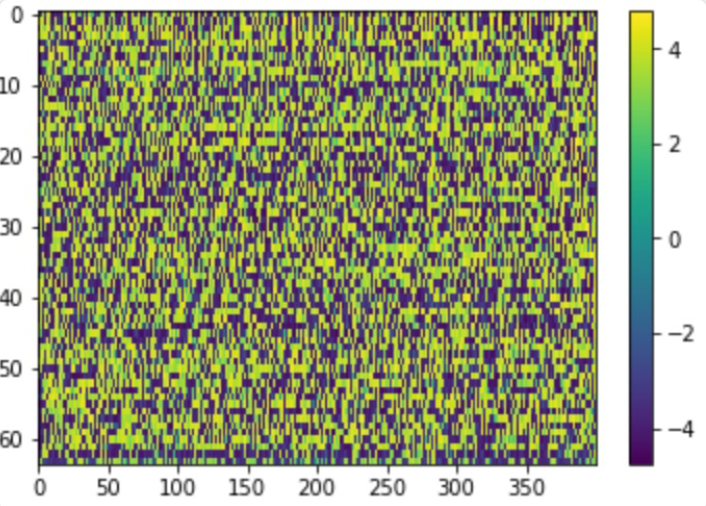}}
\centering
\caption{$\ell_{\infty}$ Attack $\epsilon =0.3$}
\end{subfigure}
\caption{Adversarial Noises constrained with different temporal/filter banks masks.}
\label{fig:spec_noise}
\vspace{-0.3cm}
\end{figure}

To identify the robust/non-robust regions in the audio features, we compute constrained adversarial noises as is shown in Fig.~\ref{fig:spec_noise} (a,b,c). Against the best CSN audio-only model (without fusion), we computed different permutation of adversarial perturbations that mask out different frequencies and temporal portions with different attack strength. The results are shown in Table.~\ref{tab:specadv} and Table.~\ref{tab:tempadv}.
As we can observe, adversarial noises present on lower frequencies tend to have higher attack potency. This suggests that the lower frequency bands contribute more to the overall performance of the model while the features there are not robust. Attacks on higher frequencies tend to be less effective.  
Not surprisingly, higher $\epsilon$ leads to more effective attack. 
Interestingly, using the $\ell_{\infty}$ attack, attacks on the different frequency lead to similar accuracy drop. This is because $\ell_{\infty}$ is too potent as is shown in Fig.~\ref{fig:spec_noise} (d).\footnote{Due to space constraint, we often discuss adversarial perturbation using $\ell_{2}$ norm instead of $\ell_{\infty}$ norm in this work. We have performed the equivalent experiments using $\ell_{\infty}$ norm, and observed the same trend.}In the temporal domain, we can observe nearly uniform drop in performance despite of the different temporal masks, suggesting the temporal domain contributes equally to robustness.

\begin{table}[h]
\centering
\begin{tabular}{@{}llll | rrr@{}}
\toprule
 freq mask  & $\epsilon$ & norm  & $\alpha$  &{\bf mAP} & {\bf AUC} & {\bf d-prime}  \\ \midrule
 No  & -   & - &  -   & 0.392 & 0.967 & 2.598     \\
 No & 0.1 & 2 & 0.01 & 0.262 & 0.942 & 2.218      \\
 No & 0.3 & 2 & 0.01 & 0.104 & 0.865 & 1.558     \\
 0-20 & 0.1 & 2 & 0.01 & \bf{0.192} & 0.910 & 1.900     \\
 0-20 & 0.3 & 2 & 0.01 & \bf{0.077} & 0.827 & 1.334     \\
 20-40 & 0.1 & 2 & 0.01 & 0.223 & 0.927 & 2.061    \\
 20-40 & 0.3 & 2 & 0.01 & 0.084 & 0.835 & 1.376     \\
 40-64 & 0.1 & 2 & 0.01 & 0.266 & 0.942 & 2.217     \\
 40-64 & 0.3 & 2 & 0.01 & 0.121 & 0.871 & 1.602     \\
\bottomrule
\end{tabular}
        \caption{Performance of our best performing CSN audio-only model under the attack of constrained adversarial perturbations in different \emph{frequency} domains. }
        \label{tab:specadv}
\end{table}

\begin{table}[h]
\centering
\begin{tabular}{@{}llll | rrr@{}}
\toprule
 temporal mask  & $\epsilon$ & norm  & $\alpha$  &{\bf mAP} & {\bf AUC} & {\bf d-prime}  \\ \midrule
 No  & -   & - &  -   & 0.392 & 0.967 & 2.598     \\
 0-200 & 0.1 & 2 & 0.01 & 0.274 & 0.945 & 2.266     \\
 0-200 & 0.3 & 2 & 0.01 & 0.171 & 0.902 & 1.830     \\
 200-400 & 0.1 & 2 & 0.01 & 0.272 & 0.945 & 2.266     \\
 200-400 & 0.3 & 2 & 0.01 & 0.169 & 0.903 & 1.835     \\
\bottomrule
\end{tabular}
        \caption{Performance of our best performing CSN audio-only model under the attack of constrained adversarial perturbations in different \emph{temporal} domains. }
        \label{tab:tempadv} 
\vspace{-0.5cm}
\end{table}


\subsection{Adversarial Robustness of different Neural Architectures}
\label{sec:archi}

We also compute the adversarial noise with the same $\ell_2$ norm and $\epsilon=0.3$ to attack different neural network models.
We run 3 trials of every model with different seeds to reported the average performance.
The results are shown in table \ref{tab:results}\footnote{To avoid repeating, refer to Table~\ref{tab:fusion} for early/middle fusion result}. We observe that replacing the recurrent layer of CRNN with transformers greatly boost the audio classification performance, suggesting that self-attention modules could lead to best accuracy empirically.
Under the influence of adversarial noise, our full late fusion model still has the highest performance, whereas ResNet seems to suffer the least in terms of performance drop. This is an interesting yet not fully understood phenomena that pure convolutional modules appear to be more robust than recurrent or self-attention layers. We conjecture that the residual links in ResNet could be obfuscating the gradients making them more robust against gradient based attacks. 

\begin{table}[h]
\setlength\tabcolsep{3.2pt}
\begin{center}
    \caption{Different architectures against the same adversarial noise} \label{tab:results}
    \begin{tabular}{c c c c | c c c c} 
    \toprule
    \multirow{2}{*}{\bf Model} & \multicolumn{3}{l|}{\bf No attack}& \multicolumn{3}{l}{\bf Attack} \\
    & { mAP} & {AUC} & {d-prime}  &  {mAP} & {AUC} & {d-prime}\\
    \midrule
    ResNet(audio) & 0.352  & 0.966 & 2.602  & 0.214  & 0.873 & 1.412\\
    CRNN(audio)  & 0.356 & 0.966 & 2.572  & 0.183 & 0.810 & 1.123 \\
    CSN(audio) & 0.392 & 0.970 & 2.650  & 0.221 & 0.872 & 1.213 \\
    CSN+3DCNN & 0.441 & 0.969 & 2.631  & 0.271 & 0.910 & 1.871\\
    \bottomrule
    \end{tabular}
\end{center}
\vspace{-0.75cm}
\end{table}

\subsection{Class-wise Performance under attack}
As a further step towards understanding what every model is learning and their reaction under the impact of adversarial noises. We further compare the performances of CSN (audio) and CRNN on different classes. Some of the classes that have large discrepancy are shown here. Although CRNN's overall performance is much lower than CSN, it outperform CSN on several classes such as \textit{Mains hum} by a large margin. From Table.~\ref{tab:classwise}, we can see that CSN has better adversarial robustness overall compared to CRNN. It is interesting to notice that classes trained with more data (the AudioSet is unbalanced) such as speech and music are relatively robust against attacks as is shown in Table~\ref{tab:top10}, and the more well-defined structured sound such as sine wave does not get affected much by adversarial noise. Plus, the higher frequency classes' accuracy tend to drop significantly with adversarial noise.


\begin{table}[h]
\begin{center}
    \caption{Class-wise mAP comparison between CRNN and CSN (audio) w/ and w/o attack. The perturbation we used in this experiment is $\ell_2$ with $\epsilon= 0.1$. mAP-adv is the performance with attack.
} \label{tab:classwise}
    \begin{tabular}{c c c | c c} 
    \toprule
    \multirow{2}{*}{\bf Class} & \multicolumn{2}{l|}{\bf CRNN}& \multicolumn{2}{l}{\bf CSN} \\
    {\bf } & {mAP} & {mAP$-$adv} & {mAP} & {mAP$-$adv} \\
    \midrule
    Mosquito & 54.1 & 6.2                        & 61.2 & 11.4\\
    Whale vocal & 45.4 & 1.8              & 33.9 & 0.5\\
    Light engine  & 37.6 & 1.2   & 49.4 & 0.3\\
    Sewing machine & 45.7 & 1.0                  & 60.3 & 1.3\\
    Sizzle & 58.6 & 0.6                          & 71.4 & 2.2\\
    Reverberation & 23.4 & 15.6                  & 16.6 & 13.9\\
    Static & 39.5 & 0.6                          & 44.2 & 1.8\\
    Mains hum & 37.8 & 0.4                       & 24.1 & 0.4\\
    \bottomrule
    \end{tabular}
\end{center}
\end{table}
\vspace{-0.5cm}

\begin{table}[h]
\begin{center}
    \caption{Top 10 Class-wise mAP comparison on the entire test dataset w/ and w/o adversarial noise of the best-performing CSN (audio) model. The perturbation we used in this experiment is $\ell_2$ with $\epsilon= 0.1$.  ``yes" indicates the performance under the influence of adversarial noise, and ``no" indicates the original performance. Some classes can still stay on the top-10 list with the attack.} \label{tab:top10}
    \begin{tabular}{lll|lll} 
    \toprule
    \multicolumn{3}{l|}{\bf Top 10 classes w/o attack}& \multicolumn{3}{l}{\bf Top 10 classes w/ attack} \\
    \midrule
    
    {\bf Attack} & {no} & {yes} & {\bf Attack}& {yes} & {no} \\
    \midrule
    EmergVehicle   & 86.9 & 50.6  & Speech           & 60.5 & 76.5\\
    ChangeRinging      & 86.5 & 0.2   & SmokeDetec    & 55.3 & 66.5\\
    Crowing             & 86.3 & 19.6  & Music            & 54.7 & 80.5\\
    CivilSiren & 85.1 & 4.3            & Plop            & 54.0 & 71.6\\
    Siren               & 84.9 & 41.6  & EmergVehicle & 50.6 & 86.9\\
    Bagpipes            & 84.8 & 26.0  & Ambulance    & 48.2 & 70.0\\
    Music               & 80.5 & 54.7  & PoliceCar    & 48.1 & 58.1\\
    Croak               & 78.9 & 13.7  & HeartSounds     & 45.4 & 59.0\\
    Speech              & 76.5 & 60.5  & SineWave        & 41.6 & 42.2\\
    \bottomrule
    \end{tabular}
\end{center}
\vspace{-0.75cm}
\end{table}

\section{Conclusion}
In this work, we addressed several important yet unexplored questions about the multi-modal adversarial robustness of neural networks in audio/visual event classification. Our work is based on the best state-of-the-art multimodal model on the Google AudioSet with 44.1 mAP. Our observations include: 1) Later fusion has better adversarial robustness compare to early fusion, 2) self-attention layers offers more accuracy but not more robustness, while convolution layers are more robust, 3) lower frequency features contribute more to the accuracy compare to high frequency features, and yet they are not robust against adversarial noises. Our findings provide some empirical evidences that could suggest better design in balancing the robustness/accuracy trade-off in multimodal machine learning.

\newpage



\bibliographystyle{IEEEbib}
\bibliography{refs}

\begin{thebibliography}{10}

\bibitem{gemmeke2017audio}
Jort~F Gemmeke, Daniel~PW Ellis, Dylan Freedman, Aren Jansen, Wade Lawrence,
  R~Channing Moore, Manoj Plakal, and Marvin Ritter,
\newblock ``Audio set: An ontology and human-labeled dataset for audio
  events,''
\newblock in {\em 2017 IEEE International Conference on Acoustics, Speech and
  Signal Processing (ICASSP)}. IEEE, 2017, pp. 776--780.

\bibitem{huang2018multimodal}
Po-Yao Huang, Junwei Liang, Jean-Baptiste Lamare, and Alexander~G Hauptmann,
\newblock ``Multimodal filtering of social media for temporal monitoring and
  event analysis,''
\newblock in {\em Proceedings of the 2018 ACM on International Conference on
  Multimedia Retrieval}. ACM, 2018, pp. 450--457.

\bibitem{wang2018polyphonic}
Yun Wang,
\newblock ``Polyphonic sound event detection with weak labeling,''
\newblock 2018.

\bibitem{vggish}
Shawn Hershey, Sourish Chaudhuri, Daniel P.~W. Ellis, Jort~F. Gemmeke, Aren
  Jansen, Channing Moore, Manoj Plakal, Devin Platt, Rif~A. Saurous, Bryan
  Seybold, Malcolm Slaney, Ron Weiss, and Kevin Wilson,
\newblock ``Cnn architectures for large-scale audio classification,''
\newblock in {\em International Conference on Acoustics, Speech and Signal
  Processing (ICASSP)}. 2017.

\bibitem{kong2019panns}
Qiuqiang Kong, Yin Cao, Turab Iqbal, Yuxuan Wang, Wenwu Wang, and Mark~D
  Plumbley,
\newblock ``Panns: Large-scale pretrained audio neural networks for audio
  pattern recognition,''
\newblock {\em arXiv preprint arXiv:1912.10211}, 2019.

\bibitem{wang2019comparison}
Yun Wang, Juncheng Li, and Florian Metze,
\newblock ``A comparison of five multiple instance learning pooling functions
  for sound event detection with weak labeling,''
\newblock in {\em ICASSP 2019-2019 IEEE International Conference on Acoustics,
  Speech and Signal Processing (ICASSP)}. IEEE, 2019, pp. 31--35.

\bibitem{kong2018audio}
Qiuqiang Kong, Yong Xu, Wenwu Wang, and Mark~D Plumbley,
\newblock ``Audio set classification with attention model: A probabilistic
  perspective,''
\newblock in {\em 2018 IEEE International Conference on Acoustics, Speech and
  Signal Processing (ICASSP)}. IEEE, 2018, pp. 316--320.

\bibitem{Mesaros2016_MDPI}
Annamaria Mesaros, Toni Heittola, and Tuomas Virtanen,
\newblock ``Metrics for polyphonic sound event detection,''
\newblock {\em Applied Sciences}, vol. 6, no. 6, pp. 162, 2016.

\bibitem{li2017comparison}
Juncheng Li, Wei Dai, Florian Metze, Shuhui Qu, and Samarjit Das,
\newblock ``A comparison of deep learning methods for environmental sound
  detection,''
\newblock in {\em 2017 IEEE International Conference on Acoustics, Speech and
  Signal Processing (ICASSP)}. IEEE, 2017, pp. 126--130.

\bibitem{alwassel2019self}
Humam Alwassel, Dhruv Mahajan, Lorenzo Torresani, Bernard Ghanem, and Du~Tran,
\newblock ``Self-supervised learning by cross-modal audio-video clustering,''
\newblock {\em arXiv preprint arXiv:1911.12667}, 2019.

\bibitem{xu2018large}
Yong Xu, Qiuqiang Kong, Wenwu Wang, and Mark~D Plumbley,
\newblock ``Large-scale weakly supervised audio classification using gated
  convolutional neural network,''
\newblock in {\em 2018 IEEE International Conference on Acoustics, Speech and
  Signal Processing (ICASSP)}. IEEE, 2018, pp. 121--125.

\bibitem{Yu2018MultilevelAM}
Changsong Yu, Karim~Said Barsim, Qiuqiang Kong, and Bin Yang,
\newblock ``Multi-level attention model for weakly supervised audio
  classification,''
\newblock {\em ArXiv}, vol. abs/1803.02353, 2018.

\bibitem{Kong2019WeaklyLA}
Qiuqiang Kong, Changsong Yu, Yinlong Xu, Turab Iqbal, Wenwu Wang, and Mark~D.
  Plumbley,
\newblock ``Weakly labelled audioset tagging with attention neural networks,''
\newblock {\em IEEE/ACM Transactions on Audio, Speech, and Language
  Processing}, vol. 27, pp. 1791--1802, 2019.

\bibitem{Ford2019}
Logan Ford, Hao Tang, François Grondin, and James Glass,
\newblock ``{A Deep Residual Network for Large-Scale Acoustic Scene
  Analysis},''
\newblock in {\em Proc. Interspeech 2019}, 2019, pp. 2568--2572.

\bibitem{Vaswani:2017:AYN:3295222.3295349}
Ashish Vaswani, Noam Shazeer, Niki Parmar, Jakob Uszkoreit, Llion Jones,
  Aidan~N. Gomez, Lukasz Kaiser, and Illia Polosukhin,
\newblock ``Attention is all you need,''
\newblock in {\em Proceedings of the 31st International Conference on Neural
  Information Processing Systems}, USA, 2017, NIPS'17, pp. 6000--6010, Curran
  Associates Inc.

\bibitem{9053609}
K.~{Miyazaki}, T.~{Komatsu}, T.~{Hayashi}, S.~{Watanabe}, T.~{Toda}, and
  K.~{Takeda},
\newblock ``Weakly-supervised sound event detection with self-attention,''
\newblock in {\em ICASSP 2020 - 2020 IEEE International Conference on
  Acoustics, Speech and Signal Processing (ICASSP)}, 2020, pp. 66--70.

\bibitem{wang2019makes}
Weiyao Wang, Du~Tran, and Matt Feiszli,
\newblock ``What makes training multi-modal networks hard?,''
\newblock {\em arXiv preprint arXiv:1905.12681}, 2019.

\bibitem{ilyas2019adversarial}
Andrew Ilyas, Shibani Santurkar, Dimitris Tsipras, Logan Engstrom, Brandon
  Tran, and Aleksander Madry,
\newblock ``Adversarial examples are not bugs, they are features,''
\newblock {\em arXiv preprint arXiv:1905.02175}, 2019.

\bibitem{li2019adversarial}
Juncheng~B. Li, Shuhui Qu, Xinjian Li, J.~Zico Kolter, and Florian Metze,
\newblock ``Adversarial music: Real world audio adversary against wake-word
  detection system,'' 2019.

\bibitem{subramanian2019}
Vinod Subramanian, Emmanouil Benetos, and Mark~B. Sandler,
\newblock ``Robustness of adversarial attacks in sound event classification,''
\newblock in {\em Proceedings of the Detection and Classification of Acoustic
  Scenes and Events 2019 Workshop (DCASE2019)}, New York University, NY, USA,
  October 2019, pp. 239--243.

\bibitem{du2020sirenattack}
Tianyu Du, Shouling Ji, Jinfeng Li, Qinchen Gu, Ting Wang, and Raheem Beyah,
\newblock ``Sirenattack: Generating adversarial audio for end-to-end acoustic
  systems,''
\newblock in {\em Proceedings of the 15th ACM Asia Conference on Computer and
  Communications Security}, 2020, pp. 357--369.

\bibitem{tran2018closer}
Du~Tran, Heng Wang, Lorenzo Torresani, Jamie Ray, Yann LeCun, and Manohar
  Paluri,
\newblock ``A closer look at spatiotemporal convolutions for action
  recognition,''
\newblock in {\em Proceedings of the IEEE conference on Computer Vision and
  Pattern Recognition}, 2018, pp. 6450--6459.

\bibitem{GhadiyaramTM19}
Deepti Ghadiyaram, Du~Tran, and Dhruv Mahajan,
\newblock ``Large-scale weakly-supervised pre-training for video action
  recognition,''
\newblock in {\em {IEEE} Conference on Computer Vision and Pattern Recognition,
  {CVPR} 2019, Long Beach, CA, USA, June 16-20, 2019}. 2019, pp. 12046--12055,
  Computer Vision Foundation / {IEEE}.

\bibitem{Park2019SpecAugmentAS}
Daniel~S. Park, William Chan, Yu~Zhang, Chung-Cheng Chiu, Barret Zoph, Ekin~D.
  Cubuk, and Quoc~V. Le,
\newblock ``Specaugment: A simple data augmentation method for automatic speech
  recognition,''
\newblock {\em ArXiv}, vol. abs/1904.08779, 2019.

\end{thebibliography}

\end{document}